%% file: main.tex
\documentclass{article} 
\usepackage{iclr2021_conference,times}

\input{math_commands.tex}

\usepackage{hyperref}
\usepackage{url}
\usepackage{graphicx}
\usepackage{caption}
\usepackage{subcaption}
\usepackage[ruled]{algorithm2e}
\usepackage{multirow} 

\title{Practical Defences Against Model Inversion Attacks for Split Neural Networks}

\author{Tom Titcombe \\
Tessella / OpenMined \\
\texttt{tom.titcombe@tessella.com} \\
\And
Adam James Hall \\
Edinburgh Napier University / OpenMined \\
\texttt{adam@openmined.org} \\
\And
Pavlos Papadopoulos \\
Edinburgh Napier University / Apheris \\
\texttt{pavlos.papadopoulos@napier.ac.uk} \\
\And
Daniele Romanini \\
OpenMined \\
\texttt{daler.romanini@gmail.com~~~}
}

\iclrfinalcopy 
\begin{document}

\maketitle

\begin{abstract}
We describe a threat model under which a split network-based federated learning system is susceptible to a model inversion attack by a malicious computational server. We demonstrate that the attack can be successfully performed with limited knowledge of the data distribution by the attacker. We propose a simple additive noise method to defend against model inversion, finding that the method can significantly reduce attack efficacy at an acceptable accuracy trade-off on MNIST. Furthermore, we show that NoPeekNN, an existing defensive method, protects different information from exposure, suggesting that a combined defence is necessary to fully protect private user data.


\end{abstract}


\section{Introduction}
\label{intro}

Training deep learning models has typically required centralised collection of large datasets, which threatens users' privacy by handing control of sensitive data to untrusted third-parties. A recently developed method to protect privacy is Federated Learning (FL), which allows users to maintain control over their data by collaboratively training models on their own devices~\citep{mcmahan2017communication, konevcny2016federated, bonawitz2019towards}. FL makes it easier to develop models in domains such as medicine, where legal requirements impose constraints on the sharing of personal data~\citep{brisimi2018federated, kaissis2020secure}. Split learning is a variant of FL in which models are split across parties (SplitNN). Comparatively, split learning reduces the computational cost to the data owner at the expense of increasing data communication between them~\citep{vepakomma2018split}.

However, maintaining control of data during training does not solve all privacy issues. As in any computational system, neural networks are susceptible to various attacks. Recent work has demonstrated that methods to extract data~\citep{fredrikson2015model, shokri2017membership, carlini2020extracting} and model parameters~\citep{wang2018stealing} from a model, or corrupt the model performance~\citep{bagdasaryan2020backdoor} are possible. The large amounts of data required to train deep networks, and their propensity to memorisation, pose a serious privacy risk to users. Several tools for collaborative, private learning are being actively developed~\citep{ryffel2018generic, he2020fedml}. While having the potential to democratise deep learning, those tools could also present new opportunities to bad actors to attack ML models at a large scale. Therefore, it is vital to understand the practical limitations of possible attacks and defences before such systems become widely adopted.

\subsection{Contributions}
\label{contributions}

This work defines a threat model for SplitNNs in which training and inference data are stolen in a FL system. We examine the practical limitations on attack efficacy, such as the amount of data available to an attacker and their prior knowledge of the target model. In particular, we extend \textit{NoPeekNN}~\citep{vepakomma2019reducing}, a method for limiting information leakage in SplitNNs. We assess \textit{NoPeekNN}'s defensive utility in the outlined attack setting.
Additionally, we introduce a simple method for protecting user data, consisting of random noise addition to the intermediate data representation of SplitNN. We also compare this method's effectiveness to NoPeekNN\footnote{Our code can be found at \texttt{https://github.com/TTitcombe/Model-Inversion-SplitNN}}.

\section{Related Work}
\label{related}

\subsection{Federated Learning and Split Neural Networks}
\label{related-splitnn}

In Federated Learning, users collaboratively train a model while keeping sensitive data on-device by sharing the results of local training~\citep{mcmahan2017communication}. In SplitNNs, each data owner only holds part of a model, which maps input data into an intermediate data representation~\citep{vepakomma2018split, vepakomma2018no}. The other model segment, which maps the intermediate data to the task output, is held by a computational server. Like FL, SplitNNs protects users' privacy by not granting third-party access to raw input data. However, a SplitNN places less computational burden on the data owners' devices, which could typically be mobile phones or other non-specialised hardware. SplitNNs have been proposed in private learning of medical tasks \citep{vepakomma2019reducing} and vertical FL \citep{ceballos2020splitnn, romanini2021pyvertical, angelou2020asymmetric}, where multiple data owners combine model segments trained on disjoint subsets of data features.

\subsection{Model Inversion Attacks and Defences}
\label{related-attacks}

Model inversion is a class of attack which attempts to recreate data fed through a predictive model, either at inference or training time~\citep{fredrikson2015model, chen2020improved, Zhang_2020_CVPR, wu2020evaluation}. It has been shown that model inversion attacks work better on earlier hidden layers of a neural network due to the increased structural similarity to input data~\citep{he2019model}. This makes SplitNNs a prime target for model inversion attacks.

NoPeekNN is a method for limiting data reconstruction in SplitNNs by minimising the distance correlation between the input data and the intermediate tensors during model training \citep{vepakomma2019reducing}. NoPeekNN optimises the model by a weighted combination of the task's loss and a distance correlation loss, which measures the similarity between the input data and the intermediate data. NoPeekNN's loss weighting is governed by a hyperparameter $\alpha \in [0, \infty]$. While NoPeekNN was shown to reduce an autoencoder's ability to reconstruct input data, it has not been applied to an adversarial model inversion attack.

Similar to this work, \citet{abuadbba2020can} applies noise to the intermediate tensors in a SplitNN to defend against model inversion attack on one-dimensional ECG data. The authors frame this defence as a differential privacy mechanism~\citep{dwork2008differential}. However, in that work, the addition of noise greatly impacts the model's accuracy for even modest epsilon values (98.9\% to roughly 90\% at $\epsilon=10$). A similar method, \textit{Shredder}, was introduced by \cite{mireshghallah2020shredder}, which adaptively generates a noise mask to minimise mutual information between input and intermediate data. 

\section{Threat model}
\label{threat}

In this work, we consider an honest-but-curious computation server and an arbitrary number of data owners who run the correct computations during training and inference. At least one party attempts to steal input data from other parties using a model inversion attack. The attack process is as follows: 1) The attackers collect a dataset of inputs (raw data) and intermediate data produced by the first model segment. 2) They train an attack model to convert the intermediate data back into raw input data. 3) They collect intermediate data produced by some data owners and run it through the trained attack model to reconstruct the raw input data. This attack is considered a ``black box" since the internal parameters of the data owner model segment are not used in the attack. We assume that the model training process has been orchestrated by a third-party and that there is only one computational server. 

We only consider the susceptibility of inference-time data to model inversion; we do not investigate the efficacy of the attack on data collected during training. The susceptibility of split learning to other attack types (e.g. membership inference \citep{shokri2017membership}, Sybil attacks~\citep{10.5555/646334.687813, kairouz2019advances}) is out of the scope of this work. Moreover, we do not investigate ``white box" model inversion attacks, which extract training data memorised by the model segments. Alternatives to the considered threat model with different parties are detailed in Appendix~\ref{app:threatmodel}.


\section{Noise Defence and Experiments}
\label{related-dp}
\label{defence}
\label{experiments}

We introduce a \textit{noise defence} in which additive Laplacian noise is applied to the intermediate data representation on the data owners' side before sending it to the computational server. Noise is drawn from a Laplace distribution parameterized by location $\mu$ and scale $b$ each time the model is used, so the data owner's model is no longer a one-to-one function. This obscures the data communicated between model segments and makes it harder for the attacker to learn the mapping from the intermediate representation to the input data. The noise is added to intermediate data only after the model is trained (as an alternative, it could also be applied during training). We describe this other approach in Appendix~\ref{app:training-noise}. The noise defence can be applied unilaterally by the data holder. This is useful in settings where a data holder does not trust the computational server.

We qualitatively compare the noise defence to NoPeekNN for defence against model inversion. Quantitatively, we compare the distance correlation between original and reconstructed images, where a more significant correlation implies a better reconstruction, in Appendix~\ref{results-defences}. We investigate the utility of NoPeekNN and the noise defence introduced in this work, both independently and as combined.

We train classifiers in combination with NoPeekNN with loss weightings $\alpha$ equal to 0.1, 0.5, 1.0, and the noise defence mechanism with noise scales $b$ equal to 0.1, 0.5, 1.0 ($\mu=0$ for all). In all experiment, we train a simple convolutional neural network on MNIST~\citep{lecun1998gradient}. We train an attack model with similar size and architecture to the classifiers on a dataset disjoint from any data used during the training phase. The classifier's and the attacker's training processes are described in detail in Appendix~\ref{app:datasets}. Furthermore, we explore how the number of data points available to train attack models and knowledge of the data structure may impact the attack success. To validate the impact of prior knowledge, we attempt to reconstruct MNIST images using an attack model trained on EMNIST~\citep{cohen2017emnist}, which is similar to MNIST but instead contains images of handwritten characters (a-z). The impact of the dataset size is detailed in Appendix~\ref{results-attack-constraints}.

\section{Results and Conclusions}
\label{conclusion}

Both the proposed noise defence mechanism and NoPeekNN do not significantly impact the classification model's accuracy. However, NoPeekNN reduces the distance correlation, as expected, as it is explicitly optimised for that. For a detailed Table of performance of both methods and their combination with different weights, refer to Appendix~\ref{results-defences}.

\begin{figure}[bth]
  \centering
    \subfloat[]{{\includegraphics[width=0.25\linewidth]{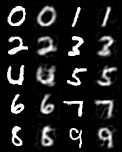} }}%
  \qquad
  \subfloat[]{{\includegraphics[width=0.25\linewidth]{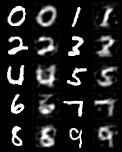} }}
  \caption{(\textbf{a}) {MNIST data reconstructions from an attacker trained on 5,000 MNIST images}. (\textbf{b}) {MNIST data reconstructions from an attacker trained on 5,000 EMNIST images}.}
  \label{fig:emnist-comparison}
\end{figure}

Figure~\ref{fig:emnist-comparison}(b) plots reconstructions of MNIST images made by an attack model trained on 5,000 EMNIST images. The reconstructions are more blurred and noisy than those made by an attack model trained on MNIST (Figure~\ref{fig:emnist-comparison}(a)), but they are still good enough to infer the class of the original images and carry some intricacies of the data points. This demonstrates that it is sufficient for an attacker to know only vague details about the dataset (``greyscale", ``simple", ``handdrawn"). In the threat model outlined in Section~\ref{threat}, the computational server could attack the data owner models without a colluding data owner. One defensive measure is to have proper and considered control of model information during training and inference: if the computational server does not need to know what the model is for, that information should be withheld. Robust identity checks could be carried out on potential data owners to ensure the computational server cannot infiltrate the system, such as the use of cryptographic identifiers proposed in \citet{papadopoulos2021privacy,abramson2020distributed}.

\begin{figure}[bth]
  \centering
    \subfloat[]{{\includegraphics[width=0.19\linewidth]{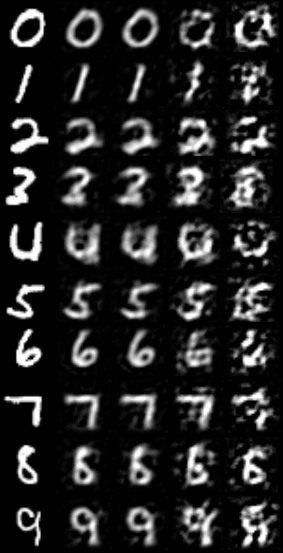} }}%
  \qquad
  \subfloat[]{{\includegraphics[width=0.15\linewidth]{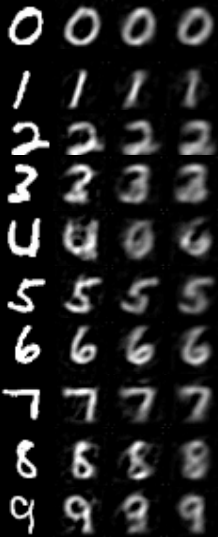} }}
  \caption{(\textbf{a}) {Model inversion attack on an MNIST classifier using the \textit{noise defence} mechanism. Left-to-right: true image, reconstructions on models with (0, 0.1, 0.5, 1.0) noise scale}. (\textbf{b}) {Model inversion attack on an MNIST classifier using NoPeekNN defence. Left-to-right: true image, reconstruction on models with (0, 0.1, 0.5) NoPeekNN weighting}.}
  \label{fig:defensive-results}
\end{figure}

Figure~\ref{fig:defensive-results} plots model inversion reconstructions of a member of each class in the dataset, applied to classifiers with varying levels of noise (Figure~\ref{fig:defensive-results}(a)) or NoPeekNN weighting (Figure~\ref{fig:defensive-results}(b)). Qualitatively, the reconstructions are completely destroyed at noise scale 1.0 (columns 5 and 10). Noise scale 0.5 (columns 4 and 9) obscures some of the more pronounced features of the input images while retaining the broad structure. Lower noise scales allow an almost complete reconstruction.
Conversely, reconstructions from NoPeekNN adopt a more generic version of the ground-truth class, but the reconstructions are more coherent. For example, slants in the digits are removed (as in classes 1, 3 and 5) as NoPeekNN weight increases.
This qualitative analysis suggests that a combination of NoPeekNN and noise defence will produce a more robust defence. Which defensive technique should be prioritised depends on the model's task. For example, a model which detects the presence of disease would prioritise class obfuscation (``Disease" or ``No Disease") using NoPeekNN. On the other hand, for facial recognition, individual user privacy is more critical, so noise should be applied to prohibit clean reconstructions.

The noise defence introduced in this work provided sufficient privacy/utility trade-offs for the MNIST dataset (Appendix \ref{results}). While simple, there are numerous tasks with a data complexity similar to MNIST, which may benefit from split learning, such as signature recognition. The privacy/utility trade-off could be improved by post-training the computational server's model segment with noise applied while keeping the data owner's model segment fixed. However, this approach removes a data owner's unilateral ability to defend against model inversion.

A Laplacian noise distribution was used in this work due to its relation to differential privacy; other noise distributions should have a similar defensive effect, and may even provide a better privacy/utility trade-off.

We have demonstrated that a user's data is susceptible to exposure by an adversary under a split neural network training setting. This happens even under limited knowledge of the task being solved. Consequently, any split learning framework where the computation server has access to the data owner model could be compromised. We have introduced a simple method for destroying data reconstructions by additive noise and shown that it protects different information about the input data from exposure than NoPeekNN.

\subsection{Future Work}
As the noise defence and NoPeekNN are introduced at the intersection between model segments, they do not protect against white box model inversion, which extracts training data memorised by a data owner's model segment. A training-time differentially private mechanism, such as DP-SGD or PATE~\citep{abadi2016deep,papernot2016semi,papernot2018scalable}, could offer protection against those attacks. Additionally, we will quantitatively investigate the privacy-utility trade-off, as explored by \cite{lecuyer2019certified}. Furthermore, the compatibility of defensive measures to protect against several possible attacks should be investigated (see Appendix \ref{limitations} for more details). In that regard, the noise defence may be formulated as a local differential privacy mechanism on the input dataset to the computational server's model part, which may confer additional defensive capability against other attacks, for example, membership inference. The more complex the dataset, the more difficult for an attack model to learn the data reconstruction mapping. The effectiveness of the noise defence in combination with NoPeekNN should be explored on more complex data sources.

\bibliography{iclr2021_conference}
\bibliographystyle{iclr2021_conference}

\appendix

\section{Threat Model Specifications}
\label{app:threatmodel}

In a split learning training scheme, a data owner controls a part of a model which maps input data into an intermediate data representation. A computational server controls a second part of the model, which maps the intermediate data into the desired output. As a single data owner may not have a large enough dataset to sufficiently train a model, multiple data owners may be employed to run federated training. In this scenario, each data owner has a copy of the same model segment. The complete model may be utilised for inference by actors not involved in the training process, in which case each model user must also hold a copy of the model segment.

The attack can be achieved by different parties under different settings: 
\begin{enumerate}
    \item \textbf{A computational server colluding with a data owner}. The data owner provides a dataset to train the attack model and the model segment. The computational server has access to intermediate data produced by any data owner on which the attack is run.
    \item \textbf{A computational server acting alone}. The computational server obtains access to a data owner's model segment through fraud (e.g. acting as a data owner to gain access to the system), coercion or theft and must construct their own dataset.
    \item \textbf{A data owner acting alone}. The data owner trains an attack model on their own data. The data owner intercepts intermediate data sent to the computational server by another data owner.
\end{enumerate}

It is possible that a computational server stores intermediate data provided to it during training and inference; hence, a successful attack involving the computational server compromises also historical uses of the model, including data owners' who contributed to only a single round of training. As the distribution of intermediate data generated during each round of training differs, the attack may not work for early-round training.

\section{In-Training Noise Defence}
\label{app:training-noise}

Applying noise to intermediate data during the training process allows the model to adapt to the randomness and produce models with higher utility. Algorithm~\ref{alg:noise-method} describes the training noise defence, in conjunction with NoPeekNN.

\begin{algorithm}[tbh]
    \DontPrintSemicolon
    laplacian noise scale $b$\\
    NoPeekNN weight $\alpha$\\
    Data owner model $f_1$ with weights $\theta_1$\\
    Computational server model $f_2$ with weights $\theta_2$\\
    Learning rates $\lambda_1, \lambda_2$\\
    \For{$epoch\leftarrow 1,2,\dots,N$}{
        \For{$inputs, targets\leftarrow dataset$}{
            $intermediate\leftarrow f_1(inputs)$\\
            $noise \sim L(0, b)$\\
            $intermediate\leftarrow intermediate + noise$\\
            $outputs \leftarrow f_2(intermediate)$\\
            $\theta_1 \leftarrow \theta_1 + \lambda_1 \frac{\partial }{\partial \theta_1} \alpha\mathcal{L}_{dcor}(inputs, intermediate) + \mathcal{L}_{task}(outputs, targets)$\\
            $\theta_2 \leftarrow \theta_2 + \lambda_2 \frac{\partial}{\partial \theta_2}
            \mathcal{L}_{task}(outputs, targets)$\\
        }
    }
    \caption{NoPeekNN with Noise Defence}
    \label{alg:noise-method}
\end{algorithm}

\section{Results and Discussion}
\label{results}

\subsection{Attack Constraints}
\label{results-attack-constraints}

\begin{figure}
     \centering
     \begin{subfigure}[b]{0.3\textwidth}
         \centering
         \includegraphics[width=\textwidth]{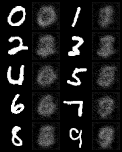}
         \caption{Data reconstructions from an attacker trained on 100 images.}
         \label{fig:100-attacker-data-points}
     \end{subfigure}
     \hfill
     \begin{subfigure}[b]{0.3\textwidth}
         \centering
         \includegraphics[width=\textwidth]{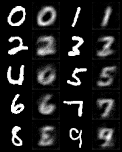}
         \caption{Data reconstructions from an attacker trained on 250 images.}
         \label{fig:250-attacker-data-points}
     \end{subfigure}
     \hfill
     \begin{subfigure}[b]{0.3\textwidth}
         \centering
         \includegraphics[width=\textwidth]{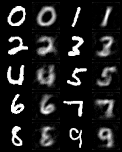}
         \caption{Data reconstructions from an attacker trained on 1250 images.}
         \label{fig:1250-attacker-data-points}
     \end{subfigure}
     \caption{MNIST images R
     reconstructions from attacks model trained on a different numbers of datapoints. The plots represents the reconstruction of images extracted from a classifier with no defence mechanisms applied. The figures show one example for each class of the MNIST dataset. Columns 1 and 3 are real datapoints; columns 2 and 4 are reconstructions.}
     \label{fig:data-point-attack}
\end{figure}

Figure~\ref{fig:data-point-attack} plots reconstructions of data extracted from an MNIST classifier by an attack model trained on different numbers of datapoints. At 250 images (Figure~\ref{fig:250-attacker-data-points}), the class of each data point can be inferred from the reconstruction, except for the examples of classes 4 and 8. However, the intricacies of each specific datapoint have not been captured in the reconstruction. The reconstructions of the attack model trained on 1250 images (Figure~\ref{fig:1250-attacker-data-points}), to contain many of the datapoints' intricacies. In some tasks, merely knowing the class of the data exposes user privacy (commonly the case in many medical tasks, such as cancer detection). In other cases (e.g. facial keypoint detection), it is necessary to reconstruct the intricacies of input data to expose private user information. 

However, in this setting, the attackers own and control the model segments. There are therefore no time constraints imposed on the attackers to develop the attack model, as to perform the attack they do not have to interact with a live system, which may expose their actions. Consequently, the poor attack performance achieved at low dataset sizes impedes an attacker, but does not stop them.

\subsection{Defences}
\label{results-defences}

\begin{figure}[!h]
  \centering
  \includegraphics[width=0.8\linewidth]{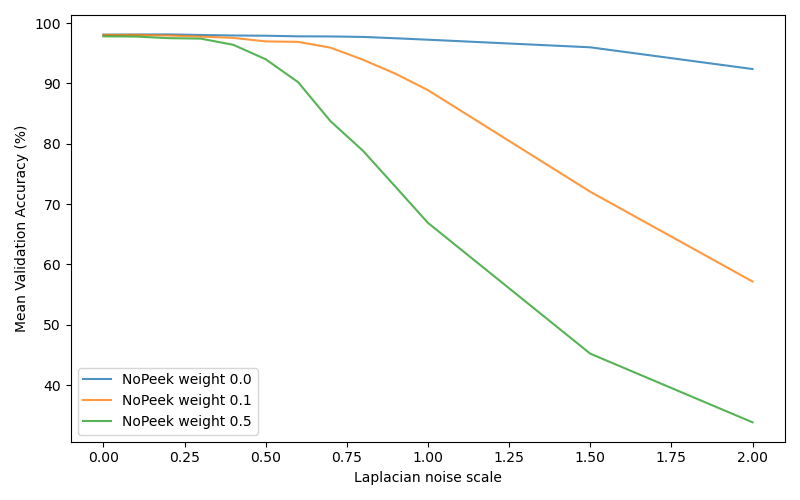}
  \caption{Accuracies on a validation dataset by classifiers with as a function of the scale of laplacian noise added to intermediate data after training.}
  \label{fig:post-train-noise-accuracies}
\end{figure}

Figure~\ref{fig:post-train-noise-accuracies} plots the accuracy of a classifier as a function of the noise scale, for noise added after training, and NoPeekNN weight. As NoPeekNN weight increases, the trade-off between noise and accuracy increases. This is likely because a greater emphasis on NoPeekNN during training produces tighter intermediate data distributions, so the same amount of noise is more abnormal and therefore destructive to the computational server model segment's capacity to model intermediate data into class probabilities. Even at high NoPeekNN's weighting loss there is a reasonable accuracy trade-off for noise scales which destroy reconstruction capability ($\approx 0.5$), which suggests that a hybrid defence would be feasible.

Table~\ref{table:training-noise-acc} shows the accuracy of classifiers on a validation dataset and the mean distance correlation between input and intermediate data. A greater NoPeekNN weight typically reduces distance correlation; this is an expected result as NoPeekNN explicitly optimises for distance correlation. Interestingly, there is a correlation between training noise scale and distance correlation, suggesting that additive noise may cancel out some of NoPeekNN's work.

\bgroup
\def\arraystretch{1.3}
\begin{table}[ht]
\centering
\begin{tabular}{cccc}
\hline
\textbf{Noise scale} & \textbf{NoPeekNN weight} & \textbf{Accuracy  (\%)} & \textbf{Distance Correlation} \\ \hline
0.0 & 0.1 & 98.04 & 0.472 $\pm$ 0.004 \\
0.0 & 0.2 & 97.80 & 0.390 $\pm$ 0.003 \\
0.0 & 0.5 & 97.90 & 0.368 $\pm$ 0.004 \\ \hline
0.1 & 0.0 & 98.19 & 0.804 $\pm$ 0.004 \\
0.1 & 0.1 & 97.84 & 0.474 $\pm$ 0.003 \\
0.1 & 0.5 & 98.00 & 0.411 $\pm$ 0.004 \\ \hline
0.2 & 0.0 & 98.00 & 0.811 $\pm$ 0.004 \\ 
0.2 & 0.1 & 97.98 & 0.491 $\pm$ 0.003 \\
0.2 & 0.5 & 97.46 & 0.411 $\pm$ 0.003 \\ \hline
0.5 & 0.0 & 98.24 & 0.795 $\pm$ 0.004 \\
0.5 & 0.1 & 97.52 & 0.525 $\pm$ 0.003 \\
0.5 & 0.5 & 97.38 & 0.437 $\pm$ 0.003 \\ \hline
\end{tabular}
\caption{Validation accuracy and distance correlation between input data and intermediate tensor of classifiers using NoPeekNN and training noise defences}
\label{table:training-noise-acc}
\end{table}
\egroup

\subsection{Limitations}
\label{limitations}

This work only considered the defensive utility of additive noise to model inversion attack on the intermediate data. However, it could be considered a local differential privacy process, which confers some protection to the computational server model segment against model inversion and membership inference attacks, among others. Additionally, this work only investigates the efficacy of model inversion attack on a 2-dimensional image dataset, MNIST. MNIST is a very simple, low-resolution greyscale dataset, therefore data reconstruction is relatively easy to perform. Future work should investigate the utility of the noise defence, in combination with NoPeekNN, on more complex datasets.

\section{Training Details}
\label{app:datasets}

The MNIST dataset is pre-split into a training set, containing 60,000 images, and a test set, containing 10,000 images. We use the first 40,000 images of the training set to train the classifier, images 40,000-45,000 to train the attacker and images 45,000-50,000 to validate the attacker. For each experiment, we train a simple convolutional neural network for 10 epochs with a batch size of 32. Larger batch sizes are limited by the computational complexity of calculating distance correlations. We use the Adam optimizer with a learning rate of 0.001 for both model segments. In the attack setting outlined in this work, attackers have access to the classifier model; therefore, usage of an attack model with an appropriate architecture for the target model is expected. The attack models are trained with the same hyperparameters. As the attack models worked ``out-the-box", a dedicated adversary with unlimited access to a target model could achieve greater attack efficacy.

\end{document}

%% file: math_commands.tex
\usepackage{amsmath,amsfonts,bm}

\def\eqref#1{equation~\ref{#1}}

\def\1{\bm{1}}

\DeclareMathAlphabet{\mathsfit}{\encodingdefault}{\sfdefault}{m}{sl}
\SetMathAlphabet{\mathsfit}{bold}{\encodingdefault}{\sfdefault}{bx}{n}

